\documentclass[11pt,a4paper]{article}
\usepackage[hyperref]{emnlp2020}
\usepackage{times}

\usepackage{latexsym,graphicx,booktabs,multirow}
\usepackage{subfigure}
\usepackage{url,algorithm2e}
\usepackage{amsmath,bm}
\usepackage{stmaryrd}

\usepackage{microtype}

\aclfinalcopy 

\def\aclpaperid{732} 

\definecolor{myorange}{RGB}{230, 159, 0}
\definecolor{myblue}{RGB}{0, 114,178}

\title{Controlled Hallucinations:\\
Learning to Generate Faithfully from Noisy Data}

\author{Katja Fillippova \\
  Google Research, Berlin, Germany \\
  {\tt katjaf@google.com} \\}

\date{}

\begin{document}
\maketitle
\begin{abstract}
Neural text generation (data- or text-to-text) demonstrates remarkable performance when training data is abundant which for many applications is not the case. 
To collect a large corpus of parallel data, heuristic rules are often used but they inevitably let noise into the data, such as phrases in the output which cannot be explained by the input. 
Consequently, models pick up on the noise and may \textit{hallucinate}--generate fluent but unsupported text. 
Our contribution is a simple but powerful technique to treat such hallucinations as a \textit{controllable aspect of the generated text}, without dismissing any input and without modifying the model architecture.
On the WikiBio corpus \cite{lebret-etal-2016-neural}, a particularly noisy dataset, we demonstrate the efficacy of the technique both in an automatic and in a human evaluation.
 
\end{abstract}

\section{Introduction}

Deep neural network-based (DNN) models have demonstrated remarkable performance on a multitude of text-to-text \cite[inter alia]{bahdanau-attention,bert-to-bert,narayan-etal-2018-dont,rush-etal-2015-neural} as well as data-to-text generation tasks \cite[inter alia]{wiseman-etal-2017-challenges,puduppully-etal-2019-data}. 
To reach high performance, DNN models require a large training corpus which is normally not readily available.
Indeed, it is rare to have a sufficiently large human-curated corpus of parallel data \cite{koehn-europarl}, and researchers have come up with heuristic rules to mine input-output pairs on a large scale \cite{hermann-teaching,rush-etal-2015-neural,narayan-etal-2018-dont}. 
No matter how powerful, DNN models are known to be sensitive to data artifacts \cite{kaushik-18-reading} and pick on the noise in the training data.

\begin{figure}[t]
\includegraphics[scale=.24]{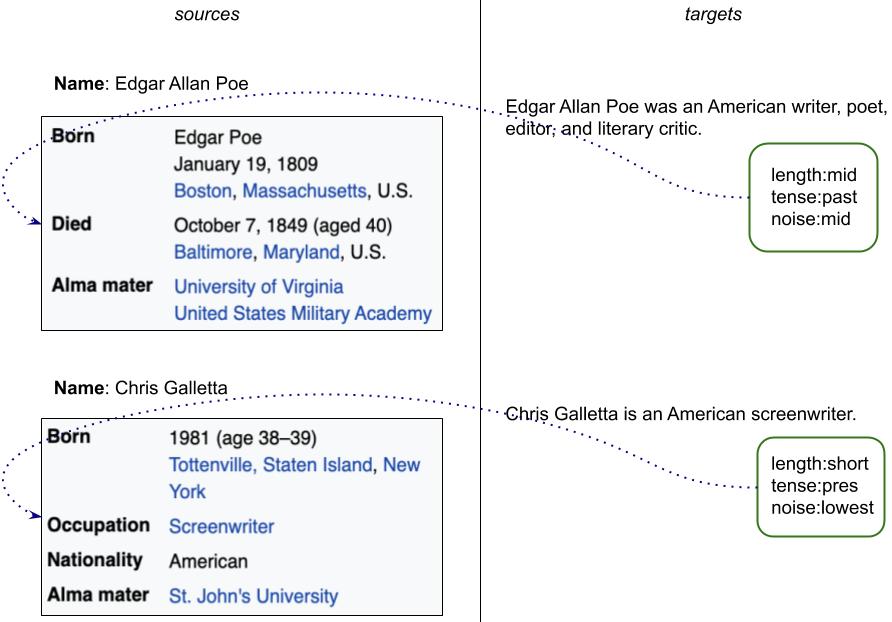}
     \caption{Two WikiBio sources and targets with example attributes: \textit{tense} and \textit{length} can be read-off the target directly. When added to the input, the model gets a knob to control for length and tense. We propose to estimate the \textit{noise} degree by comparing the source with the target thus obtaining a \textit{hallucination knob}.}
    \label{fig:knob}
\end{figure}

While \textit{hallucinations} have not been defined formally, the term is standardly used to refer to the generated content which is either unfaithful to the input, or nonsensical \cite{faithful-factual-tacl}. In our work we are concerned with the former hallucination kind which is primarily caused by imperfect quality of the training data. %
If the data are noisy, how can one reduce the chances of hallucinating?
One may try to improve the quality of a dataset and clean it from phrases for which a clear support in the input is missing, or augment the input with information found only in the output. The former path is risky as it easily results in ungrammatical targets. The latter approach of enforcing a stronger alignment between inputs and outputs has been tried previously but it assumes a moderate amount of noise in the data \cite{nie-etal-2019-simple,dusek-etal-2019-semantic}. 
Alternatively, one can leave the data as is and try to put more pressure on the decoder to pay attention to the input at every generation step \cite{tian-sticking}. This requires significant modifications to the model and may make it harder for the decoder to generate fluent and diverse text as found in the targets. 

\begin{figure}[t]
\includegraphics[scale=.36]{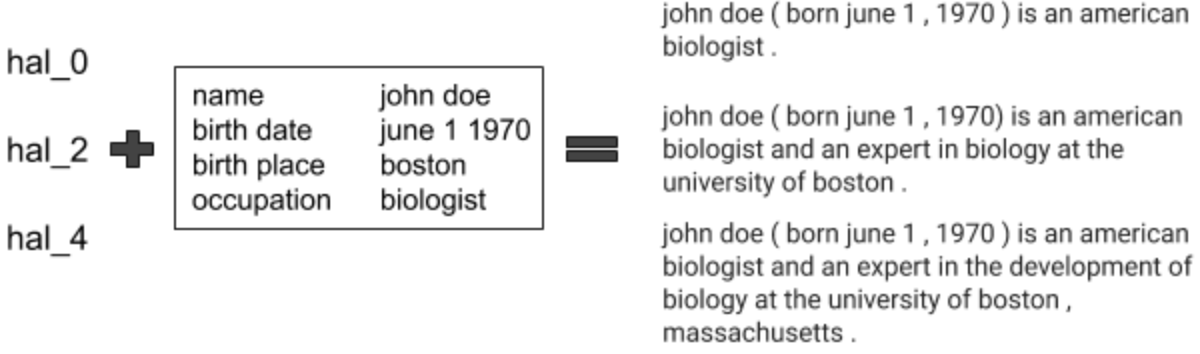}
     \caption{Example outputs from our $hal_{LM}$ model on the same input table with three hallucination degrees.}
    \label{fig:ex}
\end{figure}

In contrast to the described approaches, our proposal is to train the model on the data as is without modifying the decoding (and encoding) architecture but instead introduce \textbf{a handle} on the input side \textbf{to control} the degree of hallucination (Fig. \ref{fig:knob}). With this \textbf{"hallucination knob"} one can minimize (or maximize) the amount of unsupported information in the output during generation (Fig. \ref{fig:ex}). The hallucination or noise degree of every training instance is estimated separately and converted into a categorical value which becomes part of the input, like in a controlled generation setting \cite{ficler-goldberg-2017-controlling,raffel-etal-2019-t5}. We introduce a simple technique to measure the amount of noise in every training example which is based on the intuition that whenever a language model (LM) has a smaller loss than a conditional generator during forced-path decoding, it is a good signal that the next token cannot be explained by the input. 

We consider a particularly noisy dataset, WikiBio \cite{lebret-etal-2016-neural}, which has been found to have extra information in 62\% of the references \cite{dhingra-etal-2019-handling} and where 1:1 correspondence between the input and the output never holds \citet{perez-beltrachini-gardent-2017-analysing}. Our models demonstrate superior performance to the model of \newcite{liu-2018-structure-aware} which reports SoTA BLEU results on WikiBio. 
In sum, our contributions are (1) a novel idea of controlling hallucinations which requires no modification to the model, (2) a data- and task-independent technique of implementing this idea and (3) three-way evaluation with human raters which confirms that faithfulness does not need to be traded for coverage. 

\section{Controlling Hallucinations} 

Controlled language generation is used when when one wants the output to exhibit a certain attribute. For example, in sentence compression \cite{filippova-etal-2015-sentence} one may wish to control the length of the output to fit a length budget or fairly compare different models. This can be achieved by reading the length off the training data and using it as an additional input during training so that during inference one obtains a "length knob" \cite[Fig. \ref{fig:knob}]{kikuchi-etal-2016-controlling}. Apart from length, many other attributes like sentiment, style or theme can be controlled for, becoming an additional input for the encoder or the decoder \cite{ficler-goldberg-2017-controlling}. 
Controlled generation is a powerful technique which has recently been shown to work in a multi-task setting when the task itself becomes an attribute \cite{raffel-etal-2019-t5}.

The attribute that we are interested in controlling for is the amount of hallucinations or noise. We define a special vocabulary of \textbf{hallucination degrees} and add such a degree as a prefix to the input for every datapoint. Figure \ref{fig:ex} shows the same input prepended with three different degrees and the three corresponding outputs from our controlled model trained on WikiBio. While it is straightforward to measure output length or detect sentiment, it is less obvious how to estimate the amount of noise in a given example. In what follows, we use the words \textit{noise} and \textit{hallucinations} interchangeably. 

\section{Detecting Hallucinations in the Training Data}\label{sec:detecting}

To detect hallucinations in the training data targets, we consider (3.1) an overlap-based technique, which has a clear foundation but cannot be applied to any seq2seq task, and (3.2) a simple procedure applicable in any setting. Both methods give us a \textbf{hallucination score} $hal \in [0, 1]$ for every source-target pair. 
The scores are converted into categorical values with quantiles: five intervals, each covering 20\% of the full range, are introduced and a special tag is used for every interval. During training, the data2text model learns an embedding for each of the five tags and during inference the tag with the lowest hallucination value, \textit{hal\_0}, is used (Fig.\ \ref{fig:ex}).

\subsection{Word Overlap}

When the source and the target are similar on the token level, one can use word overlap between them to estimate how many words unsupported by the source are present in the target. More formally we define $hal$ as a function of a source-target pair (\textit{x, y}): 
\begin{align}
    hal_{WO}(x, y) = 1 - \frac{|W_{y} \cap W_{x}|}{|W_{y}|}
\end{align}
\noindent where $W$ is the set of words (in the source or the target). Note that this overlap technique only makes sense when the source and the target are in the same language and are known to be very similar. The second condition may hold to different degrees even within a dataset: for example, news publishers differ in whether they tend to write more abstractive or extractive headlines \cite{zhang-etal-2018-shaped}.

\subsection{When a LM Knows Better}

It has often been observed that hallucinations can be partially explained by a strong LM component in the decoder which tends to select the next token as a likely continuation of the sequence generated so far \cite[inter alia]{rohrbach-etal-2018-object,dusek-etal-2019-semantic}. 
This observation motivates our second method of detecting hallucinations.

Given a source and a target, how can one know if a target token $w_{y_t}$ is unsupported by the source? Consider two generation models with an identical architecture trained on the same dataset: 
\begin{itemize}
    \item $\bm{LM}$: an unconditional LM which generates the next token based on the decoded prefix and which is trained only on the targets,
    \item $\bm{LM_{x}}$: a conditional LM which is also trained to generate targets but which is additionally informed about the source.
\end{itemize}
\noindent 
On the task of generating targets from the source, during forced-path decoding, we expect $LM_{x}$ to perform better as long as the target is supported by the source because, unlike $LM$, it anticipates what may come next. For example, $LM$ will assign roughly the same probability to every month of the year while $LM_x$ will put the mass on one month, provided that the birth month is listed in the source table. 
On the contrary, whenever the next token is unexpected, it is $LM$ which reserves a small probability for it because it has been trained to predict whatever is likely to continue a given prefix, while $LM_x$ puts more probability mass on tokens related to the source. The more faithful $LM_x$, the more pronounced this difference is. 

Based on this intuition, to compute a single $hal_{LM}$ value for a source-target pair, we compute the ratio of tokens predicted incorrectly by $LM_{x}$ for which $LM$ got a smaller loss than $LM_{x}$ to the total target length $|y|$ ($w_{y_t}$ denotes the $t$'th token in the target $y$; $\tilde{w_{y_t}}$ denotes the token predicted by $LM_{x}$ at position $t$):

\begin{align}
\small
\begin{split}
    hal_{LM}(x, y) = \frac{1}{|y|} \sum^{|y|}_{t=1} \llbracket & \tilde{w_{y_t}} \neq w_{y_t}  \land \\ & p_{LM}(w_{y_t}) > p_{LM_{x}}(w_{y_t}) \rrbracket
\end{split}
\normalsize
\end{align}

For example, given a prefix \textit{first-name last-name is a}, a target \textit{first-name last-name is a french writer} and a source mentioning the profession (\textit{writer}) but not the nationality (\textit{french}), $LM_x$ will assign a high probability on the next token being the profession while $LM$ will have a small probability for any continuation, including a nationality. The smaller loss of $LM$ on the next token (\textit{french}) will signalize the presence of a hallucination. 

\section{Experiments}\label{sec:exper}

The primary goal of the experiments is to verify whether hallucinations can indeed be controlled for: we compare a seq2seq model trained on the WikiBio data as is with the same model trained with the noise attribute annotated (by the Word Overlap and LM-based methods). We also evaluate the model of \newcite{liu-2018-structure-aware},  which reported SoTA BLEU results, and the model of \citet{tian-sticking}, which was designed to generate hallucination-free output. 

In our automatic evaluation, we measure BLEU \cite{papineni-etal-2002-bleu} as well as the recently introduced PARENT metric designed specifically for data2text tasks and verified on WikiBio \cite{dhingra-etal-2019-handling}. Unlike BLEU, it compares the output not only with the reference but also measures how much of it is entailed by the input table. 

While PARENT is much more appropriate than BLEU for data2text evaluation, in its standard implementation it may miss a paraphrase of a table field in the target sentence (e.g., \textit{spouse} hardly ever occurs on the target side). It may also assign points for a match with the reference which is unsupported by the table. Thus, it can give a wrong estimate of both precision and recall and should be complemented with a human evaluation if two similar performing models are compared. 

To this end, in our experiments with human raters we measure fluency and faithfulness of generated sentences as well as coverage: we need all three as we do not want to favor models which generate fluent and faithful but short sentences because fluency and faithfulness can be trivially achieved with a handful of templates. 
\begin{description}
\item[Fluent] sentences are natural and grammatically correct (\textit{Fluent, Mostly fluent} and \textit{Not fluent}). We report the percentage of fluent sentences.
\item[Faithful] sentences express information supported by the table or by non-expert background knowledge (\textit{Faithful, Mostly faithful} and \textit{Not faithful}). Since there is a grey area of what can be inferred from the table without expert knowledge\footnote{For example, \textit{place of birth: Paris} suggests that the person is French although an exception is thinkable. \textit{position: midfielder} and \textit{club: Juventus} imply that the person is a soccer player. We observed that \textit{mostly faithful} is often used for such inferences.}, we report the percentage of Faithful and Mostly faithful sentences to the total.
\item[Coverage] counts table cells with the information expressed in the generated sentence. 
\end{description}
\noindent
Faithfulness and coverage can be seen as precision and recall metrics, respectively. We randomly selected 200 examples from the test set and collected three ratings for every input table and a generated output. 

\subsection{Model}

We train a bi-LSTM encoder-decoder model\footnote{Model details: two encoder and a single decoder layers; 256 dimensions for token embeddings, the size of the hidden cell is 128; Adam optimizer and attention; learning rate of 0.001 with a decay factor; 16,000 tokens in the vocabulary.} on WikiBio tokenized into SentencePieces \cite{kudo-2018-sentencepiece}. The input table is converted into a string with \textless row\textgreater and \textless col\textgreater special tags indicating fields and values. We use the standard train-development-test split and do no pretraining. The same model architecture is used for $LM$ and $LM_x$. That is, the default seq2seq model which we compare against is also used as $LM_x$. It differs from $LM$ in that the latter takes no input and the only difference to the controlled models is that they prepend the input with a single hallucination tag. 

\subsection{Removing Noisy Examples}

The first question we address is whether a data cleaning procedure would already result in good quality sentences. As Table \ref{tab:auto} indicates, the cleanest 20\% of the data with the smallest $hal_{WO}$ is not sufficient to train a competitive model. The predictions are more precise than those of the default model but the PARENT-recall and also the BLEU scores are low. Given a big gap to all other models, we do not evaluate this variant of the seq2seq model with humans. 

\subsection{Results}

All the models 
perform similar in terms of PARENT-F, the differences are in PARENT precision and recall. \textsc{Liu-et-al.} gets the best PARENT-F score but it comes at the cost of much lower precision than any other model which is exactly the problem we are trying to battle: unfaithful generations are arguably more harmful than missing information. Hence we turn to the human evaluation to draw final conclusions.

As perfect coverage and faithfulness can be achieved by concatenating the fields of an input table, we first verify that the generated sentences sound natural to humans. On this dimension, all the models designed to reduce hallucinations perform comparably well (93-96\%) and better than the models which do not address this problem (\textsc{Liu-et-al.}, \textsc{seq2seq}). 

Supporting the main hypothesis of our work, the two controlled versions of the seq2seq model produce significantly more faithful sentences than both \textsc{Liu-et-al.} and the default \textsc{seq2seq}: the gap to the default \textsc{seq2seq} version is 15-25 points (13-15, if mostly faithful is included). 
Contrasted with \textsc{Tian-et-al.}, our techniques are comparable or better if only faithful ratings are considered and worse if also mostly-faithful results are included.
However, \textsc{Tian-et-al.} requires significant modifications to the model (e.g., using the variational Bayes objective) which may not always be implementable. More importantly, \textsc{Tian-et-al.} is the model with the significantly smaller coverage than any other model (4.1 vs. 4.5 for $hal_{LM}$). In terms of coverage, the LM-based version of the controlled generator achieves higher coverage than the overlap-based one, equalling the default seq2seq. 

The last point is the main result of our work: it is possible to keep the recall of the default model (\textsc{seq2seq}) while dramatically improving precision. Moreover, no assumptions about the similarity between the sources and targets in the training data are needed as the $hal_{WO}$ method demonstrates.

\begin{table}[]
    \centering
    {\small
    \begin{tabular}{l|cc}
         & BLEU-4  & PARENT (P / R / F) \\
    \hline
        seq2seq (clean data)       & 31.9 & 76.3 / 37.7 / 48.1 \\
    \hline
        Liu-et-al.       & 45.4 & 74.0 / 44.0 / 52.8 \\
        Tian-et-al.             & 38.1 & 79.5 / 40.6 / 51.4 \\
    \hline
        seq2seq                & 41.0 & 75.9 / 42.0 / 51.8 \\
        seq2seq + $hal_{WO}$ & 36.5 & 79.5 / 40.9 / 51.7 \\
        seq2seq + $hal_{LM}$   & 36.1 & 78.5 / 40.3 / 50.9
    \end{tabular}
    \caption{Automatically computed metrics.}
    \label{tab:auto}
    }
\end{table}

\begin{table}[]
    \centering
    {\small
    \begin{tabular}{l|ccc}
         & Fluent  & \begin{tabular}{@{}c@{}}Faithful \\ (F+MF)\end{tabular} & Coverage \\
    \hline
        Liu-et-al.  & 89\%    & 41 (55) \% & 4.7 \\
        Tian-et-al.         & 95\%    & 68 (92) \% & 4.1 \\
        seq2seq           & 90\%    & 51 (67) \% & 4.5 \\
    \hline
        seq2seq + $hal_{WO}$ & 93\%  & 76 (82) \% & 4.3 \\
        seq2seq + $hal_{LM}$   & 96\% & 66 (80) \% & 4.5 
    \end{tabular}
    \caption{Human evaluation results.}
    \label{tab:human}
    }
\end{table}

\section{Discussion}

Comparing the two methods of estimating the amount of hallucinations in a target, for applications where the input and the output use the same vocabulary with a comparable term distribution the overlap method may be better as it has a clear foundation. 
The LM-based method that we proposed has an important advantage that it makes no assumptions about the data. In our WikiBio experiment it also produced better results in the human evaluation, presumably because it allowed for paraphrasing and straightforward inferences. For example, the target \textit{ozren nedoklan was a yugoslav footballer and manager.} has a high $hal_{WO}$ score because the source table has no occupation field and does not mention \textit{yugoslav}. The $hal_{LM}$ score of that example is zero because \textit{footballer} and \textit{manager} can be inferred from the names of the clubs and the \textit{manageryears} fields in the source.

\paragraph{Possible extensions}
It should be emphasized that alternative methods of detecting noise can be explored and may perform better in the controlled-hallucination framework. For example, it is possible to measuring target-source similarity in an embedded space or use word alignment tools to find unsupported information. 

While here we have focused on eliminating hallucinations, one can think of applications where one is interested in generating \textbf{adversarial} sentences which sound fluent but are guaranteed to include unsupported information. Figure \ref{fig:ex} shows how the amount of hallucinations in the output increases following the value of the hallucination knob.

\paragraph{Why is BLUE so different?}
It is striking that while all the models tested outperform \citet{liu-2018-structure-aware} in terms of PARENT and human evaluation scores, none could approach its BLEU performance. We do not have an explanation of why this is so but note that our results are in line with the review by \citet{reiter-2018-structured} who concludes that BLEU is an inappropriate metric for generation tasks other than MT. 

\paragraph{Can we measure length instead of noise?} One may wonder whether an even simpler approach of controlling for length would deliver a similar reduction in hallucinations. Indeed, hallucinations and length are expected to correlate, and shorter length should result in fewer hallucinations. However, as pointed out in Sec.\ \ref{sec:exper}, drastically reducing hallucinations may be possible without any control mechanism and can be achieved, at least on WikiBio, with templates. The main challenge lies in doing so without a big drop in informativeness, that is, in coverage of input fields. Comparing the outputs of $hal_{LM}$ with those of $hal_{WO}$, and both with those of \citet{tian-sticking}, we note that the ranking in terms of average sentence length (in sentencepiece tokens) coincides with the ranking in terms of coverage (Table \ref{tab:human}): 17.2, 17.8, 18.7. While $hal_{WO}$ may associate the special \textit{hal\_0} token with the shortest 20\% of the training data, for $hal_{LM}$ this token is apparently associated with a different selection of 20\% of the data points. 

\section{Conclusions}

We presented a simple but powerful idea of controlling hallucinations which are caused by the noise in the training data and proposed two ways of detecting such noise. 
We demonstrated that it is possible to reduce the amount of hallucinations at no coverage cost by informing the model about how noisy every source-target example is and without changing the model architecture. Importantly, this was done without making any assumptions about the data.
In an evaluation with humans we showed that the faithfulness of generated sentences can be significantly improved at no loss in fluency or coverage. The results we reported on the noisy WikiBio dataset improve upon the prior work.

\bibliography{paper}
\bibliographystyle{acl_natbib}

\end{document}